\documentclass[lettersize,journal]{IEEEtran}
\usepackage{amsmath,amsfonts}
\usepackage{algorithmic}
\usepackage{array}
\usepackage[caption=false]{subfig}
\usepackage{textcomp}
\usepackage{stfloats}
\usepackage{url}
\usepackage{verbatim}
\usepackage{graphicx}
\usepackage{algorithm}    
\usepackage{multirow}
\usepackage{orcidlink}
\usepackage[table]{xcolor} 
\usepackage{colortbl} 

\usepackage[numbers,sort&compress]{natbib}
\usepackage{hyperref}
\hypersetup{
    colorlinks=true,
    linkcolor=blue,
    filecolor=blue,
    urlcolor=blue,
    citecolor=blue,
}

\def\BibTeX{{\rm B\kern-.05em{\sc i\kern-.025em b}\kern-.08em
    T\kern-.1667em\lower.7ex\hbox{E}\kern-.125emX}}
\usepackage{balance}

\begin{document}
\title{SoDa$^2$: Single-Stage Open-Set Domain Adaptation via Decoupled Alignment for Cross-Scene Hyperspectral Image Classification}

\author{Yiwen Liu$^{\orcidlink{0009-0004-0545-3575}}$,
Minghua~Wang$^{\orcidlink{0000-0001-5715-130X}}$, \IEEEmembership{Member,~IEEE},
Jing~Yao$^{\orcidlink{0000-0003-1301-9758}}$, \IEEEmembership{Member,~IEEE},
Xin~Zhao$^{\orcidlink{0000-0003-0631-4628}}$, \IEEEmembership{Member,~IEEE},
Gemine~Vivone$^{\orcidlink{0000-0001-9542-0638}}$, \IEEEmembership{Senior Member,~IEEE}
\thanks{ This work is supported by the National Natural Science Foundation of China under Grants 62571271 and 62201552. This research is supported by the Natural Science Foundation of Tianjin under Grant No.24JCQNJC01890. This research is also supported by the Fundamental Research Funds for the Central Universities. (\textit{Corresponding author: Minghua~Wang.})
}
\thanks{Y. Liu, M. Wang, and X. Zhao are with the Institute of Robotics and Automatic Information System (IRAIS), College of Artificial Intelligence, and the Tianjin Key Laboratory of Intelligent Robotics (tjKLIR),  Nankai University, Tianjin 300071, China (email: liuyiwen@mail.nankai.edu.cn; wangminghua@nankai.edu.cn; zhaoxin@nankai.edu.cn).}
\thanks{J. Yao is with the Aerospace Information Research Institute, Chinese Academy of Sciences, 100094 Beijing, China (e-mail: jasonyao92@gmail.com).}
\thanks{G. Vivone is with the Institute of Methodologies for Environmental Analysis (CNR-IMAA), National Research Council, 85050 Tito, Italy (e-mail: gemine.vivone@imaa.cnr.it).}
}

\markboth{IEEE Transactions on Geoscience and Remote Sensing}%
{Decoupled Alignment for Single-Stage Open-Set Domain Adaptation in Cross-Scene Hyperspectral Image Classification}

\maketitle

\begin{abstract}
Cross-scene hyperspectral image (HSI) classification stands as a fundamental research topic in remote sensing, with extensive applications spanning various fields. Owing to the inclusion of unknown categories in the target domain and the existence of domain shift across different scenes, open-set domain adaptation techniques are commonly employed to address cross-scene HSI classification. However, existing open-set cross-scene HSI classification methods still face two critical challenges: (1) domain shift issues arising from the direct alignment of mixed spectral-spatial features; (2) high computational costs caused by two-stage training strategies. To address these issues, this paper proposes a single-stage open-set domain adaptation method with decoupled alignment (SoDa$^2$) for cross-scene HSI classification. A contribution-aware dual-modality feature extraction is customized to disentangle the characteristics from spectral sequence signals and spatial details, selectively and adaptively enhancing discriminative features. The decoupled alignment module minimizes the Maximum Mean Discrepancy (MMD) to independently reduce the spectral discrepancy and the spatial discrepancy between the source and target domains, extracting more fine-grained domain-invariant features. A cost-effective single-stage dual-branch framework is designed to learn MMD-constrainted aligned features and constraint-free intrinsic features for adaptive distinction between known and unknown classes. This framework employs a Gaussian Mixture Model (GMM) to model the squared cosine similarity distribution between the two feature types, enabling open-set recognition without prior knowledge of unknown classes. Extensive experiments on three groups of HSI datasets demonstrate that SoDa$^2$ outperforms state-of-the-art methods, achieving superior classification accuracy and model transferability for open-set cross-scene tasks. The SoDa$^2$ code will be available at \url{https://github.com/liuyiwen523/SoDa2}.
\end{abstract}


\section{Introduction}
\IEEEPARstart{H}{yperspectral} image (HSI), as one of the key technologies in remote sensing, integrates both spectral and spatial information and is widely applied in various fields such as precision agriculture, environmental monitoring, and mineral identification \cite{31ZhaZhiyuan1,33DongWenqian1,26WANGisprs2026,32ZhaZhiyuan2,27wangjstsp2025,hou2025early,10233913,peng2025newly}. Among these applications, HSI classification serves as a core task of HSI, with the primary objective of assigning a class label to each pixel based on its spectral and spatial characteristics \cite{6kumar2024deep,3011072185,34DuanPuhong1}. With the rapid advancement of deep learning, an increasing number of researchers have adopted methods such as convolutional neural network (CNN), transformer, and autoencoder (AE) to perform HSI classification more effectively \cite{7cao2020hyperspectral,8yang2022hyperspectral,9jaiswal2023integration,pang2025special,yao2022semi}.

\begin{figure}[!t]
\centering
\includegraphics[width=0.9\columnwidth]{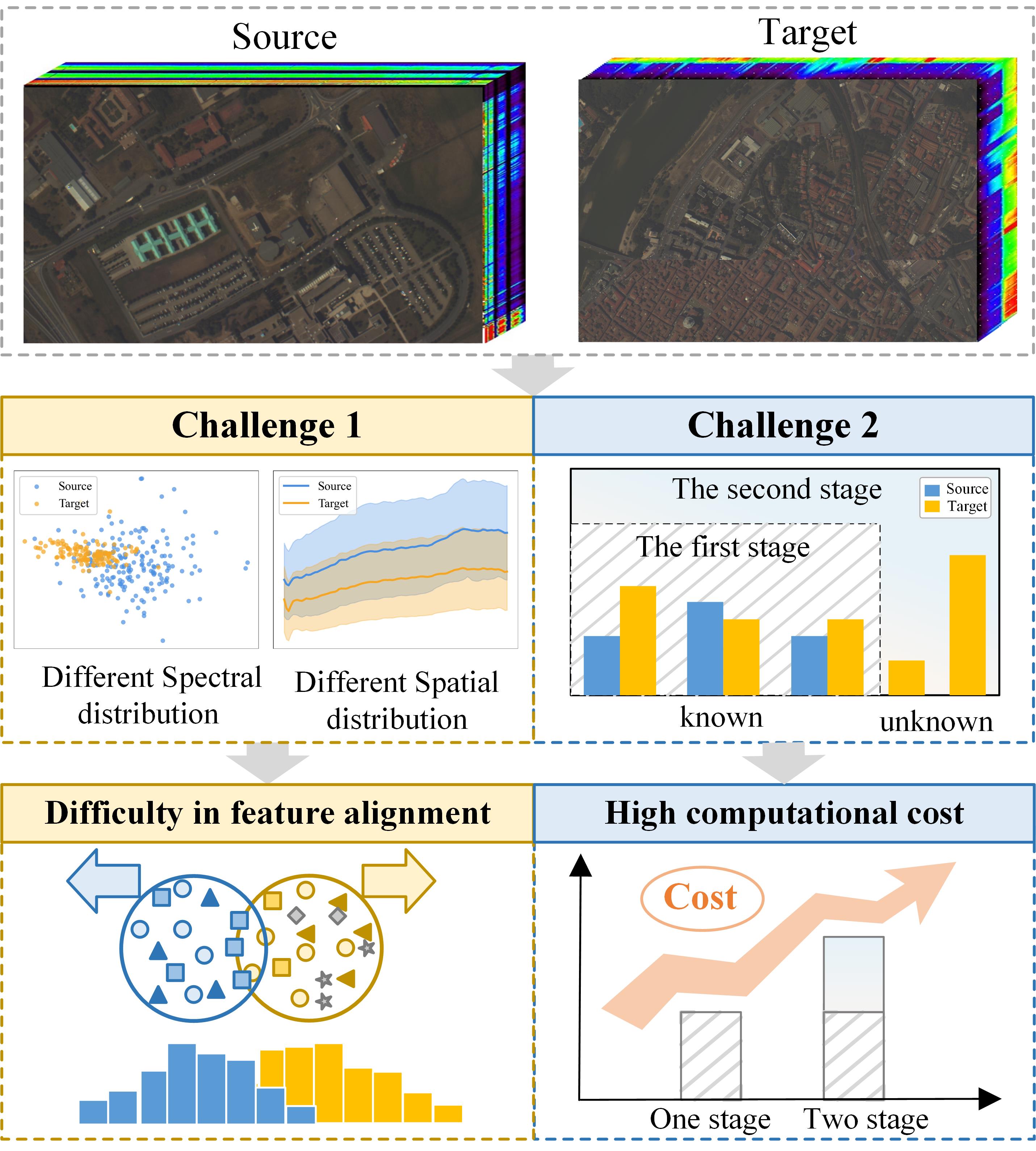}
\caption{Key Challenges of OSDA in HSI.}
\label{fig1}
\end{figure}

Although the above methods achieve remarkable performance, they rely on the assumption that training and testing datasets are independently and identically distributed \cite{109991175}. In real-world applications, however, HSIs collected at different times or locations often exhibit distribution discrepancies due to variations in sensor parameters, imaging conditions, and atmospheric effects \cite{11zhang2024locality,12jiang2025cross}. This phenomenon is commonly referred to as the domain shift problem \cite{13qi2025shift}. When a model trained on a well-labeled source domain is directly applied to a target domain with domain shift, its classification accuracy typically degrades significantly \cite{3610309248}. Domain adaptation (DA), as a kind of transfer learning, is an effective method to solve the above problems \cite{149944086}. DA methods aim to reduce the distribution discrepancy between the source and target domains by learning domain-invariant features that are insensitive to domain variations \cite{15xin2024feature}. Among them, classical DA approaches based on discrepancy minimization and adversarial learning have been widely studied and successfully applied in HSI classification \cite{16li10255730,17huang2023cross}.

Most of the aforementioned DA methods focus on closed-set, where the source and target domains are assumed to share the same set of categories \cite{MTS10292696,3510054088}. However, in real applications, the target domain often contains unknown classes that do not exist in the source domain, leading to an open-set (OS) \cite{1810227336,11408294}. In this context, open-set DA (OSDA) faces two main challenges: (1) addressing the domain shift between the source and target domains; (2) accurately distinguishing known classes from unknown ones in the target domain. Currently, an increasing number of researchers have begun to explore OSDA. Saito et al. \cite{19OSBPSaito10.1007/978-3-030-01228-1_10} proposed the open set domain adaptation by backpropagation (OSBP) model, which constructs a generator and a discriminator to address domain shift through adversarial training while distinguishing known and unknown classes. Liu et al. \cite{20STAliu8953906} introduced the separate to adapt (STA) model, which employs a binary classifier to separate known and unknown classes and a multi-class classifier to classify known samples, using adversarial learning to mitigate domain shift effects. Li et al. \cite{21ANNAli10203690} proposed a causality-based framework, adjustment and alignment (ANNA), which uses a front-door adjustment module to correct biased learning in the source domain and applies decoupled causal alignment to align cross-domain distributions. Bi et al. \cite{22WGDTbi10919145} developed an OSDA method based on a weighted generative adversarial network and dynamic thresholding (WGDT), employing a domain discriminator to prevent negative transfer and a class-anchor weighting strategy to identify unknown classes effectively.

It is worth noting that most existing OSDA methods tackle the domain shift through domain adversarial strategies \cite{38zheng2024open,39zhang2024integrating,40zhang2020open}. By constructing an adversarial game between the domain discriminator and the feature extractor, the overall distribution discrepancy between the source and target domains is minimized to achieve feature alignment. Traditional methods commonly concatenate or directly mix the spectral features and spatial features to form a unified fusion feature and then performs the overall adversarial alignment, that is, hybrid alignment. Qie et al. \cite{41Qie11245618} constructed a spectral-spatial joint feature extraction structure to extract spatial features and spectral features respectively, and fuse them at the feature level to extract domain-invariant features through adversarial learning. Xin et al. \cite{42XIN2024103850} designed a transformer-based network for extracting global spectral-spatial features and designed a domain discriminator on the mixed features to achieve the extraction of domain-invariant features. Nevertheless, as illustrated in Fig. \ref{fig1}, in HSI data, the spectral feature and the spatial feature are not only distributed independently across source and target domains but also contribute differently to the final classification results \cite{379924236}. When the inter-domain discrepancy is substantial, spectral information tend to easily distinguish classes. In contrast, when spectral features exhibit high similarity, such as asphalt and asphalt pavement, spatial detail features become more critical. Therefore, directly aligning mixed spectral-spatial features in a unified space via conventional adversarial training or similar approaches may undermine the cross-domain transferability and classification accuracy of the model.

\begin{figure}[t]
\centering
\includegraphics[width=0.95\columnwidth]{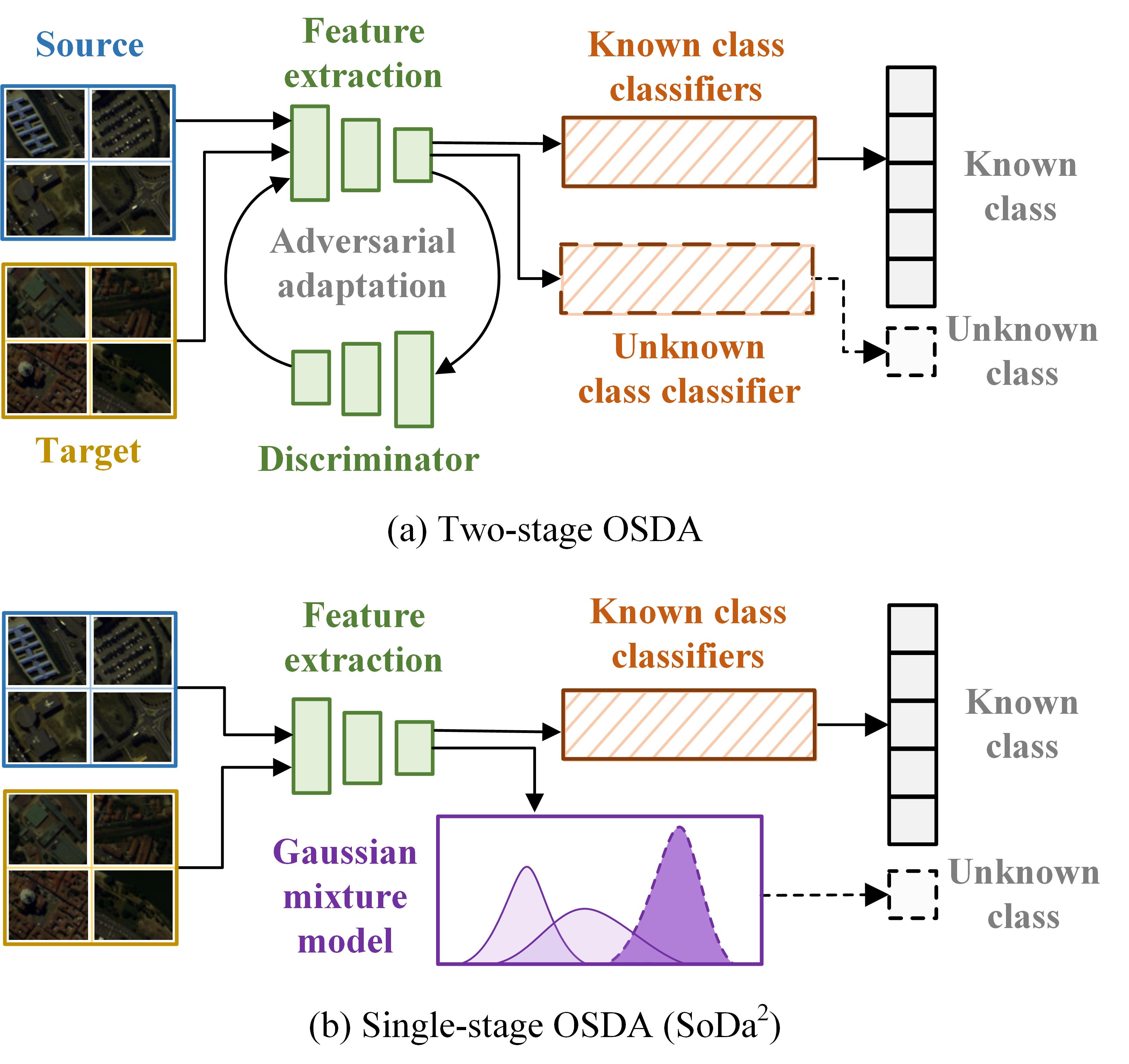}
\caption{Comparison of Open Set Domain Adaptive Models.}
\label{fig2}
\end{figure}

In addition, most of the existing OSDA methods adopt a dual-classifier architecture combined with a two-stage training strategy, as shown in Fig. \ref{fig2}. These approaches typically involve two classifiers: one for known classes and another for unknown classes. In the first stage (the pre-training phase), the known-class classifier is trained using source domain data. In the second stage, both source and target domain data are used to train the unknown-class classifier. Although this dual-classifier, two-stage training framework has attracted considerable attention due to its strong performance, it still suffers from complex training procedures and high computational costs.

To address the aforementioned challenges, this paper proposes a single-stage open-set domain adaptation method based on decoupled alignment (SoDa$^2$) for HSI classification. The main contributions are summarized as follows:

(1) We propose a SoDa$^2$ method for HSI classification to address the challenges of domain shift and unknown classes under open-set cross-scene conditions. A contribution-aware dual-modality feature extraction and decoupled alignment are constructed to collaboratively extract independent domain-invariant features from spectral sequences and spatial details. A single-stage training framework is designed to adaptively distinguish between known and unknown classes.

(2) To learn more independent domain-invariant features, we design a dedicated decoupled alignment module that separately aligns spectral and spatial features across different scenes. Specifically, the proposed method minimizes the Maximum Mean Discrepancy (MMD) between the source and target domains along the spectral and spatial dimensions, thereby effectively alleviating domain shift in HSI classification and reducing cross-domain feature distribution discrepancies in a targeted manner.

(3) We design a single-stage training framework based on dual-branch feature extraction, which adaptively distinguishes between known and unknown classes in a cost-effective manner. This framework facilitates the learning of aligned features constrained via MMD and intrinsic features free from alignment constraints. To achieve effective separation of known and unknown classes without prior knowledge of the latter, a Gaussian Mixture Model (GMM) is employed to model the distribution of squared cosine similarity values between the aligned and intrinsic features of each target sample. 

The remainder of this article is organized as follows. Section \ref{Proposed Method} elaborates our proposed method SoDa$^2$. Section \ref{EXPERIMENTS} reports the experiments in detail and discusses the experimental results. Finally, a brief conclusion is drawn in Section \ref{CONCLUSION}.

\begin{figure*}[ht]
\centering
\includegraphics[width=0.95\textwidth]{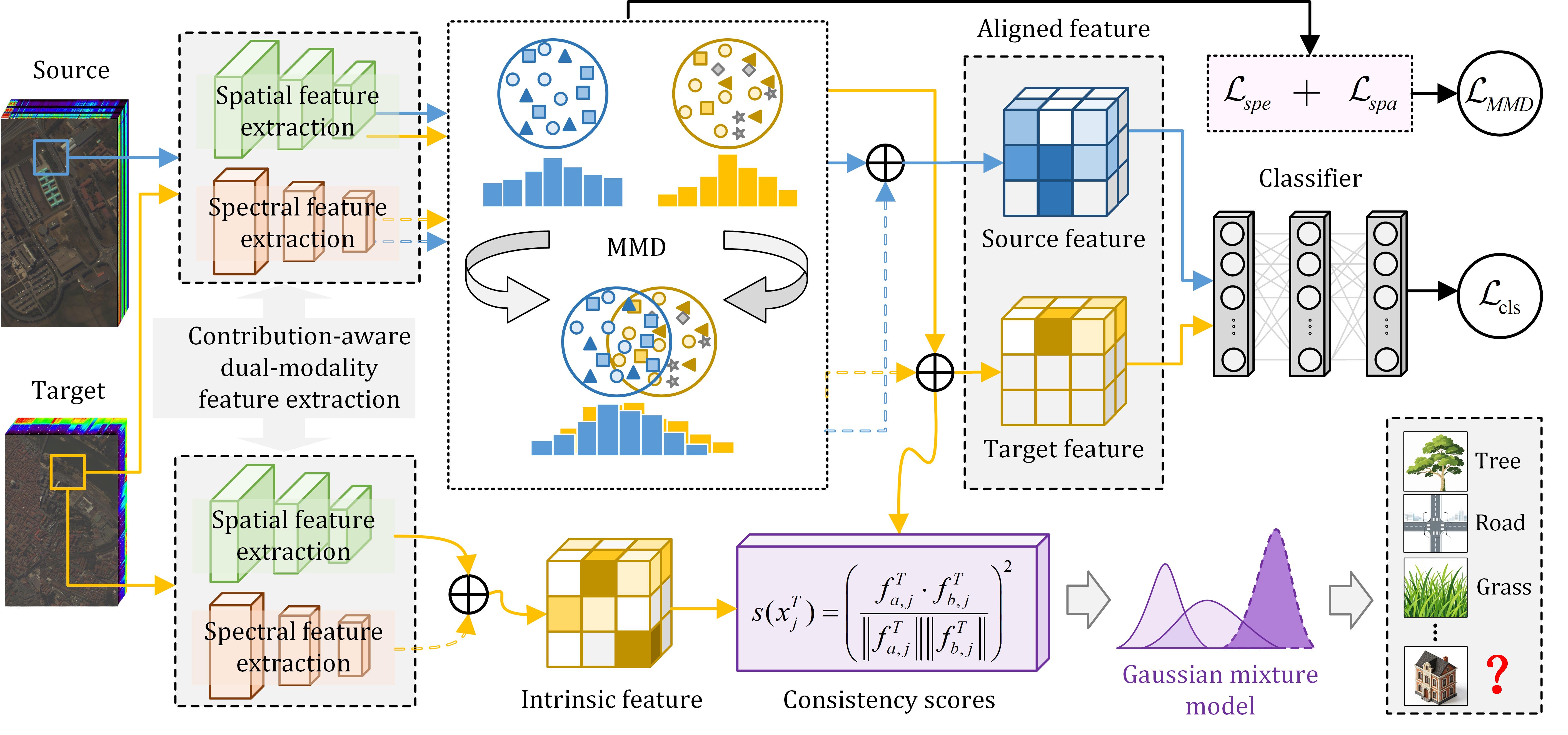}
\caption{Framework diagram of a single-stage open-set domain adaptation method based on decoupled alignment (SoDa$^2$).}
\label{fig6}
\end{figure*}

\section{Proposed Method}
\label{Proposed Method}
The source domain data are defined as ${{D}^{S}}=\{({{X}^{S}},{{Y}^{S}})\}=\{(x_{i}^{S},y_{i}^{S})\}_{i=1}^{{{n}^{S}}}$, where ${{n}^{S}}$ denotes the number of samples in the source domain. The source domain data follow the distribution $P$, and the number of categories in the source domain is ${{C}^{S}}$. The target domain data are defined as ${{D}^{T}}=\{{{X}^{T}}\}=\{x_{j}^{T}\}_{j=1}^{{{n}^{T}}}$, where ${{n}^{T}}$ represents the number of samples in the target domain. The target domain data follow the distribution $Q$, where $Q\ne P$. The number of categories in the target domain is ${{C}^{T}}$, and it can be expressed as ${{C}^{T}}={{C}^{S}}+{{C}^{Unk}}$, where ${{C}^{Unk}}$ denotes the number of unknown classes.

Our goal is to train a model that can accurately classify target domain samples belonging to the shared categories ${{C}^{S}}$, while simultaneously identifying target domain samples that belong to the unknown categories ${{C}^{Unk}}$. To achieve this, we design the SoDa$^2$. First, the contribution-aware dual-modality feature extraction is developed to separately learn the spectral and spatial contribution-aware features of each sample. Next, a decoupled alignment module minimizes the feature distribution discrepancy between the source and target domains for the shared categories. Finally, an open-set recognition module, based on the GMM, identifies known and unknown classes, with the known classes being further classified by a classifier to complete the final classification task. The overall framework of the model is illustrated in Fig. \ref{fig6}.

\subsection{Contribution-Aware Dual-Modality Feature Extraction}
\begin{figure*}[t]
\centering
\includegraphics[width=0.87\textwidth]{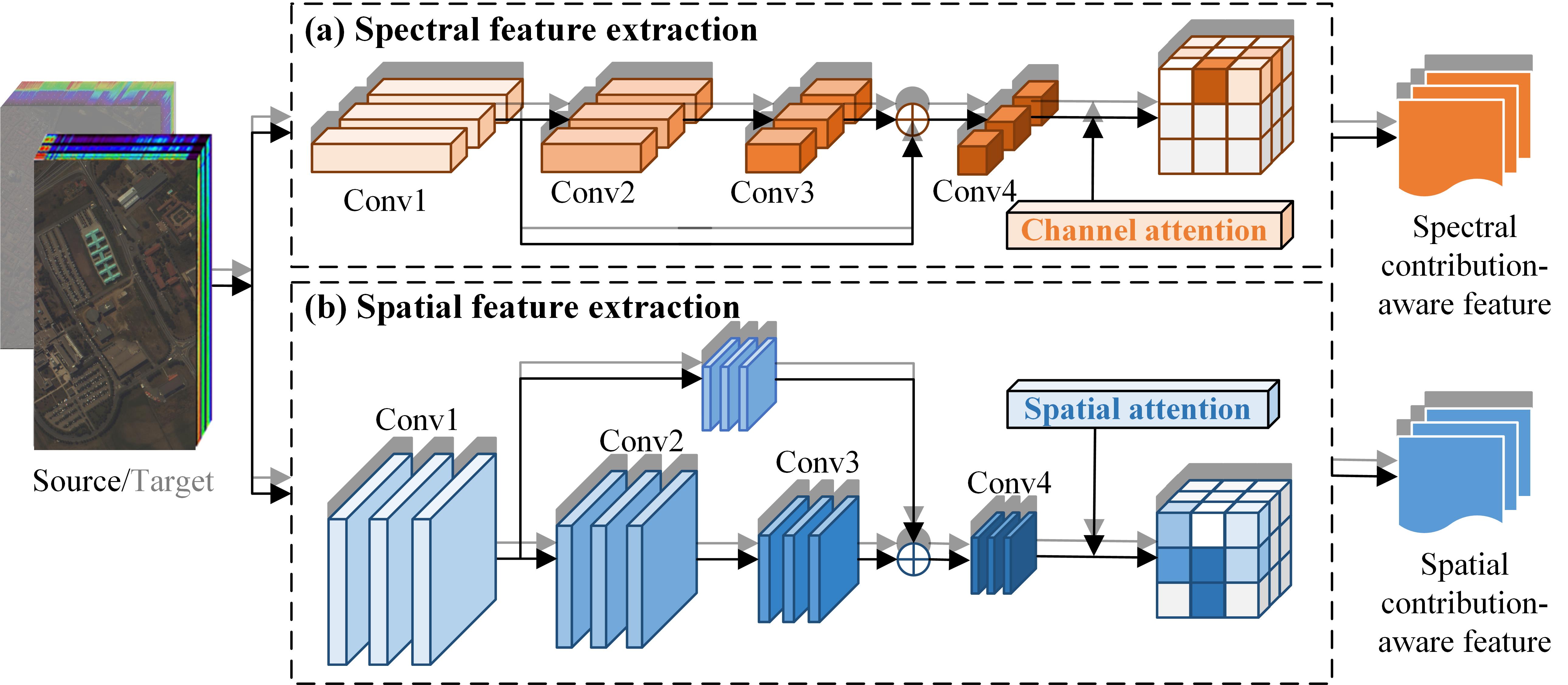}
\caption{Framework diagram of contribution-aware dual-modality feature extraction.}
\label{fig4}
\end{figure*}

Considering that a single modality feature is insufficient to comprehensively represent the characteristics of ground objects, we design a contribution-aware dual-modality feature extraction. This module consists of two parts: spectral feature extraction and spatial feature extraction, which operate in parallel to extract both spectral and spatial information, as illustrated in Fig. \ref{fig4}.

In the spectral feature extraction module, the input data first pass through an initial convolutional layer to obtain basic representations, and are then processed by two subsequent convolutional layers to progressively capture higher-order spectral information. To simultaneously preserve fine-grained band information and high-level spectral features, a cross-layer residual connection is established between the outputs of the first and third convolutional layers. After fusing the features from these two layers, the result is fed into the final convolutional layer to obtain the spectral features, which can be expressed as Equation (\ref{eq1}):
\begin{equation}
\label{eq1}
{{F}_{spe}}={{H}_{fuse}}(Conv(X),Con{{v}^{3}}(X)),
\end{equation}
where $X$ represents the input data, ${{F}_{spe}}$ represents the extracted spectral features, ${{H}_{fuse}}$ denotes the residual fusion operation, $Conv(\cdot )$ represents the convolution operation, and $Con{{v}^{3}}$ indicates three successive convolutional layers.

In the spatial feature extraction module, the model employs three 3D convolutional layers to extract spatial information from the data, enabling the capture of global spatial structures and local texture features. To enhance the contribution of shallow spatial information to deep features, the output of the first convolutional layer is used as a residual branch, which is then processed through an additional convolutional transformation. This residual branch is then fused with the output of the third convolutional layer to ensure that shallow local details are preserved within the deep spatial features. The fused spatial features are further processed by the fourth convolutional layer to obtain the final spatial representation, which can be expressed as Equation (\ref{eq2}):
\begin{equation}
\label{eq2}
{{F}_{spa}}={{H}_{fuse}}(Con{{v}^{2}}(X),Con{{v}^{3}}(X)),
\end{equation}
where ${{F}_{spa}}$ denotes the extracted spatial features, and $Con{{v}^{2}}(X)$ represents the residual features obtained after two convolutional operations.

HSI contains massive amounts of information, including a certain portion of redundant data. To select the spectral bands and spatial details that contribute significantly to the classification task, this model introduces an attention mechanism to extract contribution-aware features. The spectral features are processed through a channel attention module, which adaptively reweights each channel according to its importance, thereby emphasizing the most informative spectral information. The spatial features are processed through a spatial attention module, which assigns higher weights to important spatial locations to strengthen the representation of key regions. The spectral and spatial contribution-aware feature after attention enhancement can be expressed as Equation (\ref{eq3}):
\begin{equation}
\label{eq3}
{\tilde{F}_{spe}}={{A}_{c}}({{F}_{spe}}),{\tilde{F}_{spa}}={{A}_{s}}({{F}_{spa}}),
\end{equation}
where ${\tilde{F}_{spe}}$ represents the spectral contribution-aware feature of the channel attention mechanism ${{A}_{c}}(\cdot)$, ${\tilde{F}_{spa}}$ represents the spatial contribution-aware features after the spatial attention mechanism ${{A}_{s}}(\cdot)$.

\subsection{Decoupled Alignment}
To overcome performance degradation caused by distribution discrepancies between the source and target domains, we design a decoupled alignment module. The core idea of this module is to independently align spectral and spatial contribution-aware features at the feature level, thereby enhancing the generalization ability and classification accuracy of the model on target-domain samples. Traditional adversarial DA methods typically align features in a coarse-grained manner. Such global alignment strategies struggle to simultaneously accommodate the heterogeneity and independently distributed nature of spectral and spatial modalities, which may lead to cross-modal feature confusion. Therefore, our module adopts a discrepancy-based domain adaptation approach to reduce feature distribution differences.

Common discrepancy-based methods include Kullback-Leibler (KL) divergence \cite{KLji2020kullback}, MMD \cite{MMDarticle}, and Correlation Alignment (CORAL) \cite{CORALsun2016deep}. KL divergence relies on accurate estimation of probability density functions for both domains, which is difficult to achieve in high-dimensional feature spaces. CORAL aligns distributions by matching second-order statistics, yet linear second-order metrics are often insufficient for describing the complex nonlinear properties of hyperspectral features. In contrast, MMD employs kernel mappings within a reproducing kernel hilbert space to flexibly capture nonlinear high-dimensional distributions. This makes MMD particularly well-suited for cross-domain alignment in hyperspectral tasks. Accordingly, this study imposes separate MMD constraints on the spectral and spatial contribution-aware features to reduce the distribution discrepancy between the source and target domains within their respective feature subspaces. The decoupled alignment loss for spectral contribution-aware features is formulated as Equation (\ref{eq4}):
\begin{equation}
\label{eq4}
{{\mathcal{L}}_{{spe}}}=\left\|\frac{1}{|\tilde{F}_{spe}^{S}|}\sum\limits_{{{f}^{S}}\in \tilde{F}_{spe}^{S}}{\phi}({{f}^{S}})-\frac{1}{|\tilde{F}_{spe}^{T}|}\sum\limits_{{{f}^{T}}\in \tilde{F}_{spe}^{T}}{\phi}({{f}^{T}})\right\|_{H}^{2},
\end{equation}
where ${\phi(\cdot)}$ denotes the nonlinear feature mapping into the Reproducing Kernel Hilbert Space (RKHS), and $H$ represents the RKHS. Similarly, the decoupled alignment loss for spatial contribution-aware features is computed as Equation (\ref{eq5}):
\begin{equation}
\label{eq5}
{{\mathcal{L}}_{{spa}}}=\left\|\frac{1}{|\tilde{F}_{spa}^{S}|}\sum\limits_{{{f}^{S}}\in \tilde{F}_{spa}^{S}}{\phi}({{f}^{S}})-\frac{1}{|\tilde{F}_{spa}^{T}|}\sum\limits_{{{f}^{T}}\in \tilde{F}_{spa}^{T}}{\phi}({{f}^{T}})\right\|_{H}^{2}.
\end{equation}

Therefore, the overall decoupled alignment loss is defined as Equation (\ref{eq6}):
\begin{equation}
\label{eq6}
{{\mathcal{L}}_{MMD}}={{\mathcal{L}}_{spe}}+{{\mathcal{L}}_{{spa}}}.
\end{equation}

After achieving decoupled alignment, the spatial and spectral contribution-aware features are concatenated along the channel dimension to obtain fused domain-invariant features.

\subsection{Open Set Recognition Module}
To effectively distinguish known classes from unknown classes in the target domain, we design an open-set recognition module. The core idea of this module is to construct two parallel feature-extraction branches and quantify the consistency between the representations generated for the same target sample. By measuring the semantic agreement between these two representations, the module enables reliable identification of unknown-class samples.

\begin{figure}[t]
\centering
\includegraphics[width=1\columnwidth]{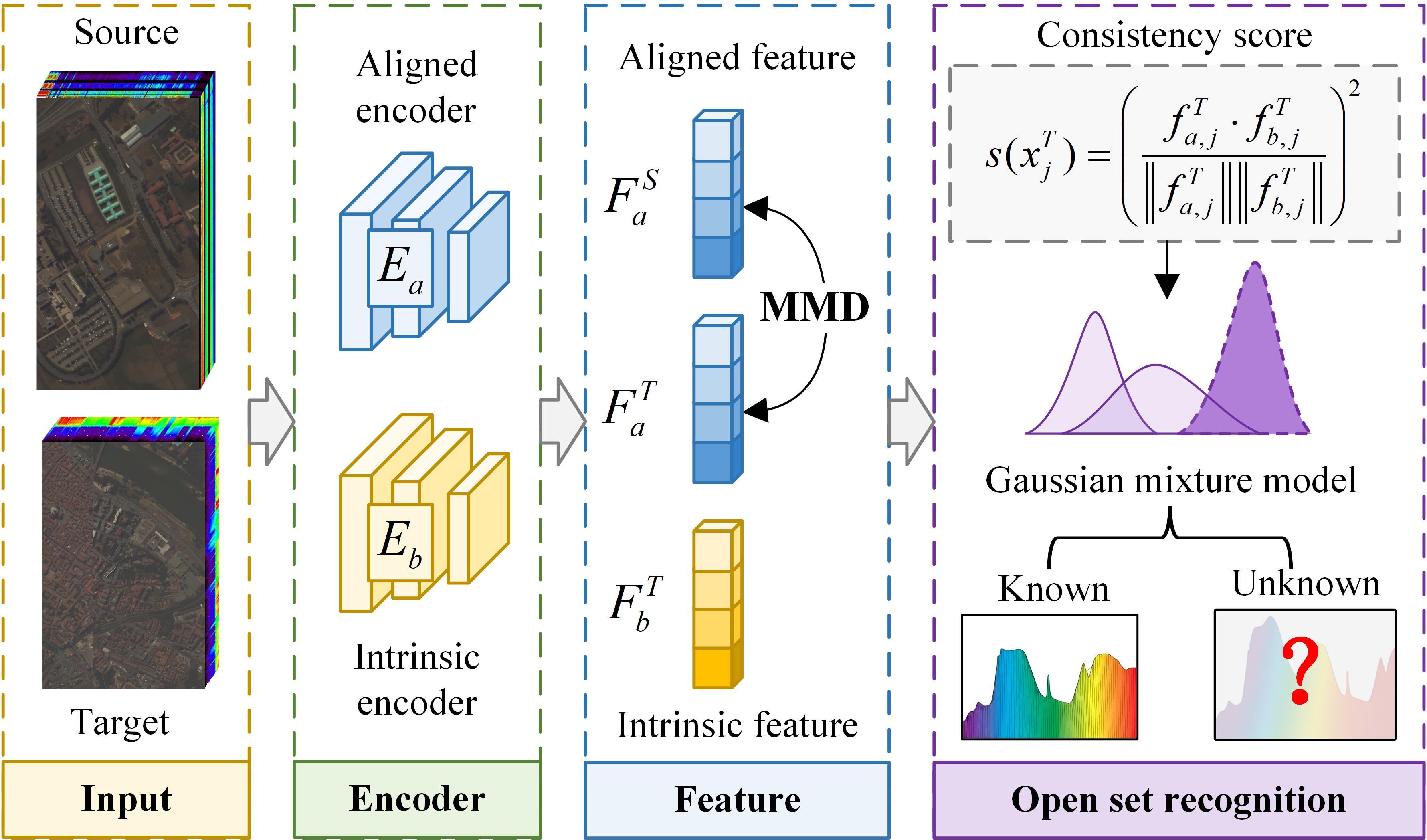}
\caption{Open set recognition module.}
\label{fig5}
\end{figure}

As illustrated in Fig. \ref{fig5}, the open-set recognition module consists of two branches: an aligned feature encoder and an intrinsic feature encoder. The aligned feature encoder aims to capture semantic information shared between the source and target domains for known classes. Its key mechanism is the introduction of MMD as a cross-domain alignment constraint. By minimizing the feature distribution discrepancy between domains in the RKHS, this branch is encouraged to learn domain-invariant representations. For a target-domain sample ${{X}^{T}}$, its aligned feature is formulated as Equation (\ref{eq7}):
\begin{equation}
\label{eq7}
F_{a}^{T}={{E}_{a}}({{X}^{T}};{{\theta }_{a}}),
\end{equation}
where ${{F}_{a}^{T}}$ denotes the aligned feature of the target sample, ${{E}_{a}}(\cdot )$ represents the aligned feature encoder—comprising a spectral feature extractor and a spatial feature extractor, and ${{\theta }_{a}}$ is its parameter set.

In contrast, the intrinsic feature branch operates without any cross-domain constraints and is designed to capture target-domain intrinsic information, with special emphasis on discriminative characteristics of unknown classes. This branch relies solely on target-domain data for optimization, thus avoiding the potential suppression of target-specific information caused by domain-alignment constraints. The intrinsic feature is given by Equation (\ref{eq8}):
\begin{equation}
\label{eq8}
F_{b}^{T}={{E}_{b}}({{X}^{T}};{{\theta }_{b}}),
\end{equation}
where $F_{b}^{T}$ denotes the intrinsic feature, ${{E}_{b}}(\cdot )$ is the intrinsic feature encoder (sharing the same architecture as ${E}_{a}$ but with independent parameters), and ${{\theta }_{b}}$ is its parameter set.

To measure the semantic correlation of feature representations from the two branches, the module employs cosine similarity as the consistency metric. For a target sample $x_{j}^{T}$, the cosine similarity between its aligned feature $f_{a,j}^{T}\in F_{a}^{T}$ and intrinsic feature $f_{b,j}^{T}\in F_{b}^{T}$ is computed as Equation (\ref{eq9}):
\begin{equation}
\label{eq9}
sim(x_{j}^{T})=\frac{f_{a,j}^{T}\cdot f_{b,j}^{T}}{\left\| f_{a,j}^{T} \right\|\left\| f_{b,j}^{T} \right\|}.
\end{equation}

This value reflects the directional alignment of the two feature vectors. In an open-set setting, known-class samples are dominated by shared cross-domain semantics due to the decoupled alignment constraint, resulting in low consistency between aligned and intrinsic features. In contrast, unknown-class samples lack source-domain supervision, making intrinsic features the primary descriptors of their semantic content. Consequently, they tend to exhibit higher directional consistency with aligned features. To enhance discriminability and ensure non-negativity, we square the cosine similarity to obtain the final consistency score by Equation (\ref{eq10}):
\begin{equation}
\label{eq10}
s(x_{j}^{T})={{\left( \frac{f_{a,j}^{T}\cdot f_{b,j}^{T}}{\left\| f_{a,j}^{T} \right\|\left\| f_{b,j}^{T} \right\|} \right)}^{2}}.
\end{equation}

Based on the consistency scores $s({{X}^{T}})$, this method assumes that all target-domain samples follow a mixture distribution composed of $K$ gaussian components. These components represent two types of samples: those belonging to known classes, which generally exhibit lower consistency scores, and those belonging to unknown classes, which usually present higher scores. The distribution of the scores is modeled by a GMM, expressed as Equation (\ref{eq11}):
\begin{equation}
\label{eq11}
p(s)=\sum\limits_{k=1}^{K}{{{\pi }_{k}}}\mathcal{N}(s\mid {{\mu }_{k}},\sigma _{k}^{2}),
\end{equation}
where ${{\pi }_{k}}$, ${{\mu }_{k}}$, and ${\sigma _{k}}$ denote the mixing coefficient, mean, and variance of the $k$-th gaussian component, respectively. 

After fitting the model using the Expectation–Maximization algorithm, samples are classified by comparing the means of the gaussian components. Samples assigned to the component with the largest mean are identified as unknown, while the remaining samples are considered known. The decision rule is written as Equation (\ref{eq12}):
\begin{equation}
\label{eq12}
\mathcal{D}({{x}_{t}})=\left\{ \begin{array}{*{35}{l}}
   \text{Unknown,} & \text{if }\arg {{\max }_{k}}{{\gamma }_{k}}(s(x_{j}^{T}))=\arg {{\max }_{k}}{{\mu }_{k}}  \\
   \text{Known,} & \text{otherwise}  \\
\end{array} \right.,
\end{equation}
where ${{\gamma }_{k}}(s(x_{j}^{T}))$ is the posterior probability that the score $s(x_{j}^{T})$ belongs to the $k$-th gaussian component. 

This probabilistic mechanism adaptively captures the distribution patterns of consistency scores and avoids the need for explicitly defining unknown-class characteristics, thereby enabling automatic open-set sample recognition without prior knowledge.

\begin{algorithm*}[t]
\caption{The specific procedure of SoDa$^2$}
\label{alg1}
{\textbf{Input:}} Source domain data ${{D}^{S}}=\{({{X}^{S}},{{Y}^{S}})\}$, Target domain data ${{D}^{T}}=\{{{X}^{T}}\}$, Decoupled alignment loss weight $\alpha$, Number of Gaussian components $K$.

{\textbf{Output:}} Class probabilities of target-domain samples
\begin{algorithmic}[1]
\STATE Feed the source-domain data ${{X}^{S}}$ into the aligned encoder to obtain the spectral contribution-aware features $\tilde{F}_{spe,a}^{S}$ and spatial contribution-aware features $\tilde{F}_{spa,a}^{S}$ through equation (\ref{eq1}), equation (\ref{eq2}), and equation (\ref{eq3});
\STATE Feed the target-domain data ${{X}^{T}}$ into the aligned encoder to obtain the spectral contribution-aware features $\tilde{F}_{spe,a}^{T}$ and spatial contribution-aware features $\tilde{F}_{spa,a}^{T}$ through equation (\ref{eq1}), equation (\ref{eq2}), and equation (\ref{eq3});
\STATE Compute the decoupled losses for spectral and spatial contribution-aware features using equation (\ref{eq4}) and (\ref{eq5});
\STATE Obtain the total decoupled loss ${{\mathcal{L}}_{MMD}}$ as in equation (\ref{eq6});
\STATE The spectral contribution-aware feature $\tilde{F}_{spe,a}^{S}$ and spatial contribution-aware feature $\tilde{F}_{spa,a}^{S}$ are concatenated by channels to obtain the aligned feature $F_{a}^{S}$;
\STATE Input $F_{a}^{S}$ into the classifier to obtain class probabilities and calculate the classification loss ${{\mathcal{L}}_{\text{cls}}}$ using equation (\ref{eq13}); 
\STATE Calculate the overall loss ${{\mathcal{L}}_{total}}$ via equation (\ref{eq14}) and update model parameters by minimizing this loss;
\STATE Feed the target-domain data ${{X}^{T}}$ into the intrinsic encoder to extract spectral contribution-aware features $\tilde{F}_{spe,b}^{T}$ and spatial contribution-aware features $\tilde{F}_{spa,b}^{T}$;
\STATE The $\tilde{F}_{spe,a}^{T}$ and $\tilde{F}_{spa,a}^{T}$ are concatenated by channels to obtain the aligned feature $F_{a}^{T}$;
\STATE The $\tilde{F}_{spe,b}^{T}$ and $\tilde{F}_{spa,b}^{T}$ are concatenated by channels to obtain the intrinsic feature $F_{b}^{T}$;
\STATE Calculate the consistency score $s$ between the $F_{a}^{T}$ and $F_{b}^{T}$ via equation (\ref{eq10});
\STATE Fit a GMM to the consistency scores according to equation (\ref{eq11}), obtaining the means of $K$ Gaussian components;
\STATE According to equation (\ref{eq12}), identify samples belonging to unknown classes as those corresponding to high-mean components, while samples associated with lower-mean components are considered known classes;
\STATE Feed the identified known samples into the trained classifier to obtain the final class probabilities for target-domain samples.
\end{algorithmic}
\end{algorithm*}

\subsection{Loss Function}
After distinguishing known and unknown samples in the target domain through the open-set recognition module, the samples identified as belonging to known classes are further classified by a known-class classifier. This classifier is trained on the source domain data and is capable of discriminating between different known classes. It is optimized using the standard cross-entropy loss function, formulated as Equation (\ref{eq13}):
\begin{equation}
\label{eq13}
{{\mathcal{L}}_{\text{cls}}}=-\frac{1}{{{n}^{S}}}\sum\limits_{i=1}^{{{n}^{S}}}{\sum\limits_{c=1}^{{{C}^{s}}}{y_{i,c}^{S}}}\log \left( p(c|f_{i}^{S}) \right),
\end{equation}
where $y_{i,c}^{S}$ denotes the ground-truth label indicating whether the $i$-th source sample belongs to class $c$, and $p(c|f_{i}^{S})$ represents the probability that the multi-modal feature $f_{i}^{S}$ of the $i$-th source sample is assigned to class $c$.

It is worth noting that the proposed method adopts a single-stage training paradigm, integrating open-set recognition and known-class classification into a unified end-to-end optimization framework. Unlike conventional two-stage approaches, all modules in our framework are jointly optimized during training. During training, each mini-batch contains both source-domain and target-domain samples. The overall loss function is defined as Equation (\ref{eq14}):
\begin{equation}
\label{eq14}
{{\mathcal{L}}_{total}}={{\mathcal{L}}_{\text{cls}}}+\alpha {{\mathcal{L}}_{MMD}},
\end{equation}
where ${{\mathcal{L}}_{total}}$ denotes the total loss of the proposed SoDa$^2$ algorithm, and $\alpha$ is the weighting coefficient for the decoupled  alignment loss.

The specific procedure for the proposed single-stage training model SoDa$^2$ is shown in Algorithm \ref{alg1}.

\section{EXPERIMENTS}
\label{EXPERIMENTS}
In this subsection, we conducted experiments on three groups of commonly used cross-scene HSI datasets to evaluate the effectiveness of the proposed SoDa$^2$ method. First, we present a detailed introduction to the datasets, model parameters, and evaluation metrics. Subsequently, a comprehensive qualitative and quantitative comparison is made between the proposed method SoDa$^2$ and the existing state-of-the-art techniques. Finally, through a series of detailed ablation experiments and parameter analyses, we verify the role of each core module and determine the optimal parameter configuration.

\subsection{Dataset Description}
Experiments were conducted on six HSI datasets, namely Pavia University (PU), Pavia Center (PC), Houston 2013 (HU13), Houston 2018 (HU18), Ziyuan1-02D Yancheng (ZY), and GaoFen-5 Yancheng (GF). These six HSI datasets were further divided into three cross-scenario tasks: PU–PC, HU13–HU18, and ZY–GF.

(1) \textit{PU-PC}: The PU dataset was acquired by the Reflective Optics System Imaging Spectrometer (ROSIS) sensor developed by the German Aerospace Center. The imaging area covers a university in Pavia of northern Italy and the contiguous regions surrounding this academic site. The spatial resolution of PU is 1.3 meters, with a spatial dimension of 610 x 340 pixels and 103 bands included. The PC dataset is also derived from the ROSIS sensor, with its imaging area located in the central region of Pavia in northern Italy. It shares the same spatial resolution of 1.3 meters as the PU dataset, and has a spatial dimension of 1096 × 715 pixels with 102 spectral bands. During the experiment, PU was employed as the source domain dataset, while PC served as the target domain dataset. A total of 102 spectral bands shared by both PU and PC were selected as the input spectral bands. The classes of tree, asphalt, brick, bitumen, shadow, meadow and bare soil were designated as the known categories. In addition, the tiles class, is defined as the unknown class to simulate the open-set scenario.

(2) \textit{HU13-HU18}: The HU13 and HU18 datasets were derived respectively from the IEEE Geoscience and Remote Sensing Society data fusion contests held in 2013 and 2018. Both datasets cover the imaging area of the university of houston campus and its surrounding regions, with a consistent spatial resolution of 1 meter. Specifically, the HU13 dataset has an image size of 349 × 1905 pixels and encompasses 144 spectral bands, whereas the HU18 dataset features an image size of 209 × 955 pixels with 48 spectral bands. In this study, HU13 was designated as the source domain data, while HU18 was treated as the target domain data. The 48 spectral bands shared by both HU13 and HU18 were selected as the model input. In the experiment, the evergreen trees and deciduous trees in HU18 were merged into trees class. Seven categories, namely grass healthy, grass stressed, trees, water, residential buildings (RB), non-residential buildings (NRB), and Road, were defined as the known classes. The remaining 12 classes in the HU18 dataset were designated as the unknown classes.

(3) \textit{ZY-GF}: Both the ZY and GF datasets capture wetland scenes in Yancheng City, Jiangsu Province, China, with image sizes of 1398 × 942 pixels and 1175 × 585 pixels, respectively. The ZY dataset is acquired by the Ziyuan1-02D satellite, while the GF dataset is collected by the GaoFen-5 satellite. The spatial resolution of both datasets is 30 m and they both contain 147 spectral bands. In the experiment, the ZY dataset was designated as the source domain, and the GF dataset as the target domain. Architecture, Sea and Offshore water are identified as the known classes shared by the ZY and GF datasets, with the other four categories exclusive to the GF dataset being assigned to the unknown class set.

\subsection{Setup}
(1) \textit{Implementation Details}: All experiments are conducted using Python 3.8.20 with PyTorch 1.12.0. The computing environment is configured with CUDA 11.6, and all models are trained on an NVIDIA RTX A6000 GPU. During training, the stochastic gradient descent (SGD) optimizer was used, with a learning rate of 0.001, a weight decay of $1e^{-3}$, and a momentum of 0.9. The number of Gaussian mixture components in the GMM is set to K = 2.

(2) \textit{Metrics}: To evaluate the performance of different methods, we adopt three metrics: the mean accuracy of known classes (OS*), the accuracy of unknown classes (UNK), and their harmonic mean (HOS). The corresponding formulations are given as follows:

\begin{equation}
\label{eq15}
\begin{aligned}
\mathrm{OS}^* &= \frac{1}{\left|C^s\right|} \sum_{c \in C^s} \mathrm{Acc}_c, \\
\mathrm{UNK} &= \mathrm{Acc}_u, \\
\mathrm{HOS} &= \frac{2 \times \mathrm{OS}^* \times \mathrm{UNK}}{\mathrm{OS}^* + \mathrm{UNK}},
\end{aligned}
\end{equation}
where $\mathrm{Acc}_c$ represents the classification accuracy on known classes, and $\mathrm{Acc}_u$ represents the classification accuracy on unknown classes.

\subsection{Experimental Results}
To verify the effectiveness of the proposed method, we compare it with six OSDA approaches, including OSBP \cite{19OSBPSaito10.1007/978-3-030-01228-1_10}, STA \cite{20STAliu8953906}, DAMC \cite{DAMC9165955}, UADAL \cite{UADALjang2022unknown}, ANNA\cite{21ANNAli10203690}, MTS\cite{MTS10292696}, and WGDT \cite{22WGDTbi10919145}. Tables \ref{tab1}–\ref{tab3} report the experimental results of these methods on the PU–PC, HU13–HU18, and ZY–GF tasks, respectively. The evaluation metrics include OS*, UNK, and HOS, where higher values indicate better performance. In the tables, the best results are highlighted in bold, while the second-best results are underlined.

\begin{figure*}
\centering
\includegraphics[width=0.9\textwidth]{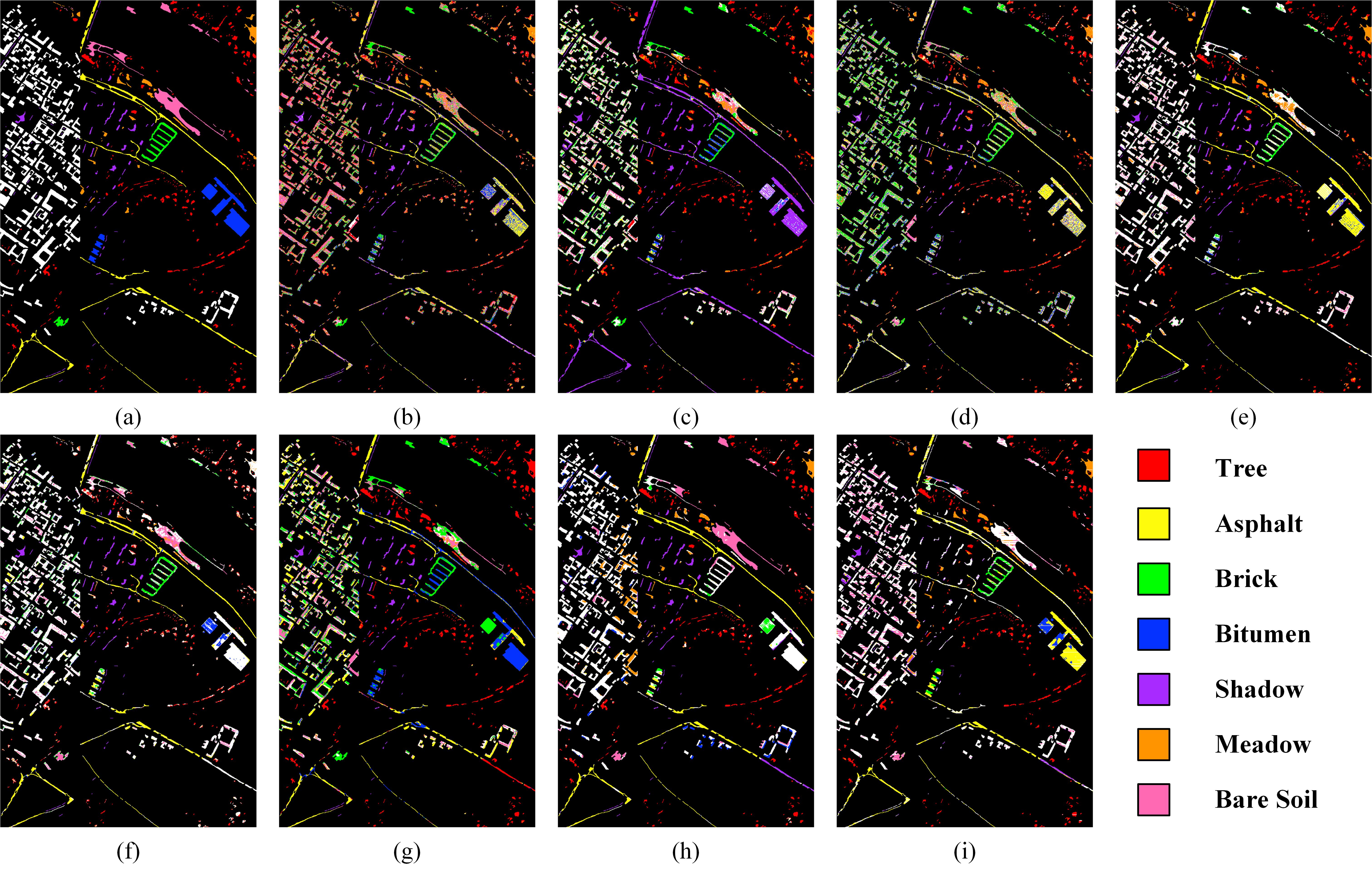}
\caption{Visualization of classification results for the PU-PC task. (a) Ground-truth. (b) OSBP. (c) STA. (d) DAMC. (e) UADAL. (f) ANNA. (g) MTS. (h) WGDT. (i) SoDa$^2$. }
\label{figr1}
\end{figure*}

Table \ref{tab1} reports the evaluation metrics of different methods on the PU–PC task. In terms of the overall metrics, the proposed SoDa$^2$ exhibits the most outstanding performance, with an HOS of 80.7\%, which is significantly higher than those of the competing methods. Regarding unknown class recognition, SoDa$^2$ attains an UNK metric of 94.3\%, outperforming all the comparison methods, which demonstrates its strong capability in identifying unknown categories under open-set scenarios. For known-class classification, although the OS* of SoDa$^2$ is only 1.6\% higher than that of WGDT, SoDa$^2$ exhibits more competitive performance across multiple categories. Among the compared methods, WGDT exhibits the second-best overall performance. UADAL and ANNA demonstrate relatively excellent performance in unknown class recognition but suffer from relatively low accuracy in known class classification. MTS shows the opposite trend, with poor performance in unknown-class recognition. OSBP and DAMC demonstrate weaker capability in identifying unknown classes, which results in comparatively lower overall metrics. Overall,  SoDa$^2$ not only achieves leading performance in both known and unknown class recognition but also maintains a favorable balance between the two aspects, which fully validates the effectiveness of the proposed method. The corresponding classification visualization results are shown in Fig. \ref{figr1}.

\begin{table*}
\centering
\caption{Experimental results of different methods on the PU-PC task} 
\renewcommand\arraystretch{1.4}
\tabcolsep=0.3cm
\begin{tabular}{c|c|ccccccc|ccc}
\hline
Method&	Venue&	Tree&	Asphalt&	Brick&	Bitumen&	Shadow&	Meadow&	Bare Soil&	OS*&	UNK&	HOS\\
\hline
OSBP \cite{19OSBPSaito10.1007/978-3-030-01228-1_10}&	ECCV'18&	47.5&	57.8&	50.8&	30.9&	92.5&	60.6&	33.7&	53.4&	9.3&	15.8\\
STA \cite{20STAliu8953906}&	CVPR'19&	80.4&	6.7&	49.6&	11.5&	98.6&	43.0&	24.8&	45.0&	53.4&	48.8\\
DAMC \cite{DAMC9165955}&	TMM'21&	71.2&	51.4&	49.2&	16.6&	90.0&	62.6&	24.2&	52.2&	25.2&	34.0\\
UADAL \cite{UADALjang2022unknown}&	NeurIPS'22&	91.3&	77.5&	44.9&	12.1&	\underline{99.8}&	\textbf{77.3}&	3.5&	58.1&	83.9&	68.6\\
ANNA \cite{21ANNAli10203690}&	CVPR'23&	57.2&	75.9&	56.9&	19.4&	98.3&	21.2&	41.2&	52.9&	86.1&	65.5\\
MTS \cite{MTS10292696}&	TCSVT'24&   \textbf{98.6}&    63.2&    \textbf{67.1}&    \textbf{67.8}&    98.5&    4.1& 32.0&  61.6&   24.8&    35.4\\
WGDT \cite{22WGDTbi10919145}&	TGRS'25&	96.1&	\underline{81.0}&	47.1&	\underline{47.0}&	\textbf{100.0}&	64.9&	\underline{46.4}&	\underline{68.9}&	\underline{86.3}&	\underline{76.6}\\
\rowcolor{blue!10}
SoDa$^2$&	\textbf{Ours}&	\underline{97.1}&	\textbf{94.0}&	\underline{66.0}&	23.5&	90.2&	\underline{75.4}&	\textbf{47.0}&	\textbf{70.5}&	\textbf{94.3}&	\textbf{80.7}\\
\hline
\end{tabular}
\label{tab1}
\end{table*}

\begin{figure*}
\centering
\includegraphics[width=0.81\textwidth]{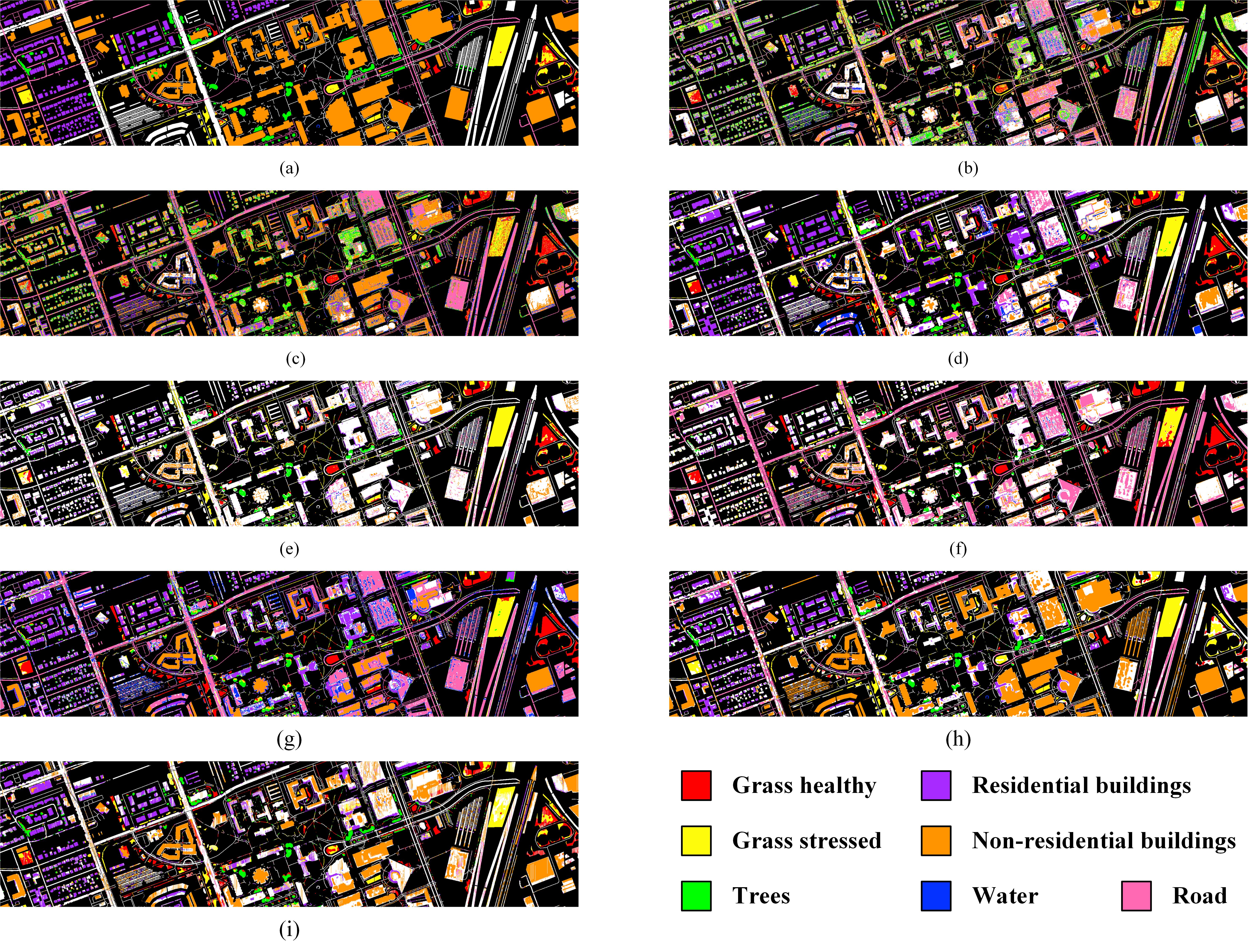}
\caption{Visualization of classification results for the HU13–HU18 task. (a) Ground-truth. (b) OSBP. (c) STA. (d) DAMC. (e) UADAL. (f) ANNA. (g) MTS. (h) WGDT. (i) SoDa$^2$.}
\label{figr2}
\end{figure*}

Table \ref{tab2} presents the experimental results of different methods on the HU13–HU18 task. The visualizations of all methods are shown in Fig. \ref{figr2}. Considering all evaluation metrics comprehensively, the proposed SoDa$^2$ method exhibits the best overall classification performance. SoDa$^2$ presents an HOS of 65.0\%, outperforming the other methods and demonstrating its comprehensive advantage in both known-class classification and unknown-class recognition tasks. For the known-class classification task, SoDa$^2$ achieves the highest OS* value of 59.0\% among all methods. From a class-wise perspective, SoDa$^2$ obtains the best classification accuracy on the Grass healthy, Grass stressed, and Residential buildings categories. Moreover, SoDa$^2$ maintains strong performance across most of the remaining categories, frequently ranking as the second-best method, which demonstrates its stability in known-class recognition. Although the UNK metric of SoDa$^2$ is 2.7\% lower than that of UADAL, the gap between the two methods is relatively small. Moreover, SoDa$^2$ demonstrates higher recognition accuracy for known classes, and when considered comprehensively, the proposed method SoDa$^2$ exhibits overall superior performance. In contrast, while UADAL attains the highest accuracy in unknown-class recognition, its performance on known-class classification is relatively limited. WGDT shows stable performance in both OS* and UNK metrics but still falls short of SoDa$^2$. Other methods, including MTS, STA, ANNA, DAMC, and OSBP, achieve competitive results in certain categories. However, their overall performance remains inferior to that of SoDa$^2$. Overall, these results demonstrate that the proposed method is highly competitive across multiple evaluation metrics. In particular, it significantly enhances unknown-class discrimination while preserving strong known-class recognition capability, thereby achieving a favorable balance and superior overall performance.

\begin{table*}
\centering
\caption{Experimental results of different methods on the HU13–HU18 task} 
\renewcommand\arraystretch{1.4}
\tabcolsep=0.3cm
\begin{tabular}{c|c|ccccccc|ccc}
\hline
Method&	Venue&	Grass healthy&	Grass stressed&	Trees&	Water&	RB&	NRB&	Road&	\textbf{OS*}&	\textbf{UNK}&	\textbf{HOS}\\
\hline
OSBP \cite{19OSBPSaito10.1007/978-3-030-01228-1_10}&	ECCV'18&	76.4&	38.1&	48.3&	64.7&	14.8&	37.2&	\underline{55.4}&	47.8&	7.0&	12.2
\\
STA \cite{20STAliu8953906}&	CVPR'19&	92.7&	51.4&	\textbf{63.6}&	30.5&	\textbf{85.0}&	11.2&	12.7&	49.6&	57.4&	53.2
\\
DAMC \cite{DAMC9165955}&	TMM'21&	66.5&	49.8&	18.3&	\underline{70.3}&	44.2&	11.2&	30.1&	41.5&	27.5&	33.1
\\
UADAL \cite{UADALjang2022unknown}&	NeurIPS'22&	91.9&	\underline{57.9}&	46.0&	1.9&	25.8&	9.0&	5.7&	34.0&	\textbf{75.0}&	46.8
\\
ANNA \cite{21ANNAli10203690}&	CVPR'23&	91.3&	29.6&	26.1&	24.4&	8.6&	7.8&	\textbf{62.9}&	35.8&	35.0&	35.4
\\
MTS \cite{MTS10292696}&	TCSVT'24& \underline{94.1}&  38.2&    46.8&   10.2&    62.8&    22.5&   49.7&    46.3&    25.2&    32.6\\
WGDT \cite{22WGDTbi10919145}&	TGRS'25&	74.5&	34.3&	48.1&	\textbf{90.6}&	65.7&	\textbf{64.7}&	3.7&	\underline{54.4}&    70.2&	\underline{61.4}
\\
\rowcolor{blue!10}
SoDa$^2$&	\textbf{Ours}&	\textbf{98.5}&	\textbf{65.5}&	\underline{60.2}&	62.0&	\textbf{85.0}&	\underline{41.3}&	0.5&	\textbf{59.0}& \underline{72.3}&	\textbf{65.0}
\\
\hline
\end{tabular}
\label{tab2}
\end{table*}

\begin{figure*}
\centering
\includegraphics[width=0.9\textwidth]{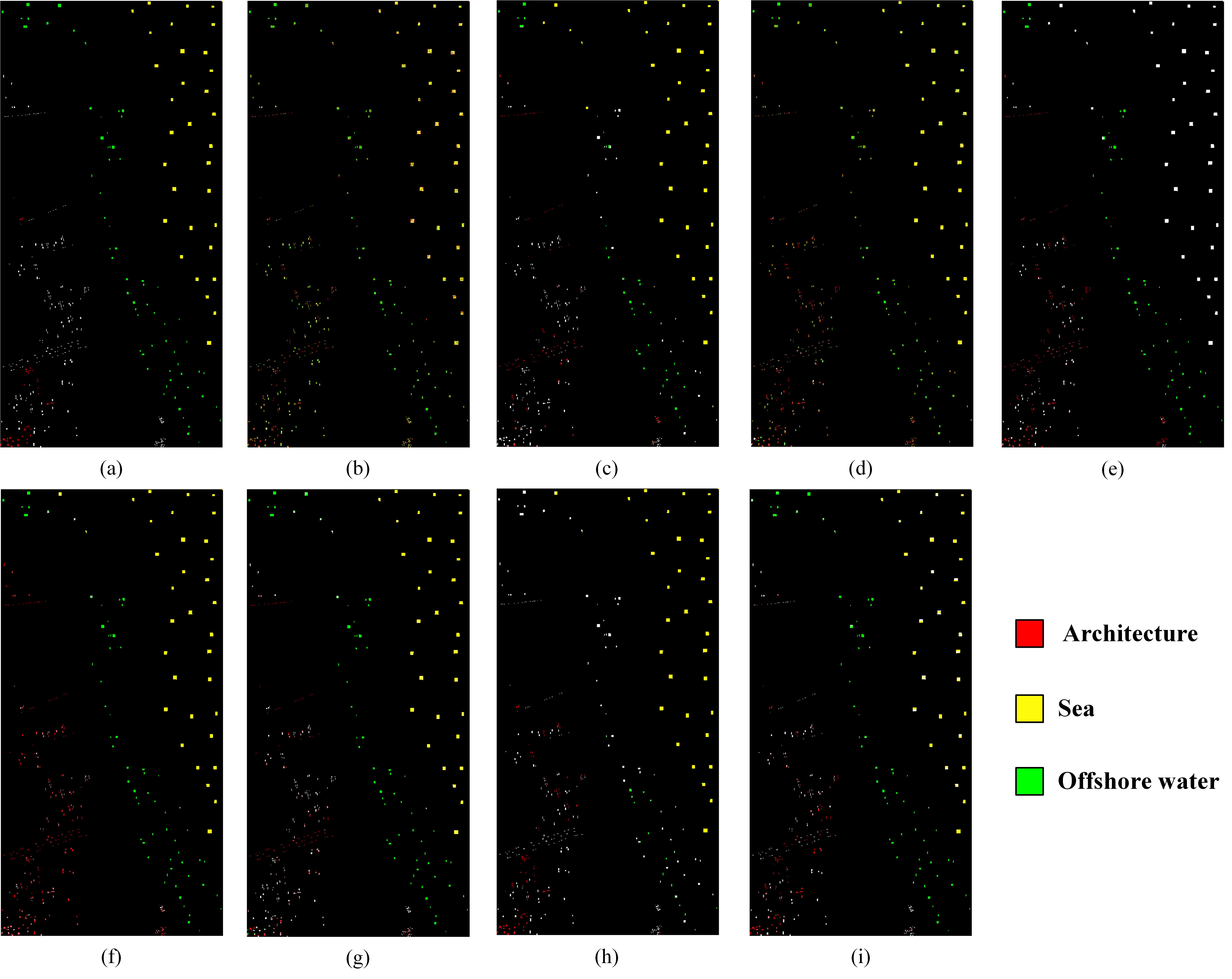}
\caption{Visualization of classification results for the ZY–GF task. (a) Ground-truth. (b) OSBP. (c) STA. (d) DAMC. (e) UADAL. (f) ANNA. (g) MTS. (h) WGDT. (i) SoDa$^2$.}
\label{figr3}
\end{figure*}

Table \ref{tab3} reports the classification performance of different methods on the ZY–GF task. Fig. \ref{figr3} shows the corresponding visualization results. Overall, SoDa$^2$ exhibits the optimal performance in terms of the comprehensive evaluation metric, with an HOS of 94.7\%, which is the highest among all the competing methods. For known-class recognition, SoDa$^2$ attains the highest OS* value of 92.1\%, indicating its discriminative capability for known categories. In the unknown-class recognition task, SoDa$^2$ also achieves the best performance, demonstrating its ability to identify unknown classes. Among the compared approaches, WGDT performs well in both known-class classification and unknown-class recognition, achieving the second-best HOS value. ANNA shows reasonable performance on known classes but suffers from relatively low accuracy in unknown-class recognition, which limits its overall performance. The STA method achieves a recognition accuracy of 100\% on the Sea category. However, its performance on the Architecture category is relatively poor. MTS also performs relatively poorly on the “Architecture” category. Other methods, such as UADAL, DAMC, and OSBP, perform relatively poorly in at least one task, whether it is known category recognition or unknown category recognition, resulting in lower overall evaluation metrics. In summary, the proposed SoDa$^2$ method not only achieves the highest accuracy in both known and unknown class recognition on the ZY–GF task but also maintains an excellent balance between the two, further validating its effectiveness in cross-source open-set HSI classification scenarios.

\begin{table*}[h]
\centering
\caption{Experimental results of different methods on the ZY–GF task} 
\renewcommand\arraystretch{1.4}
\tabcolsep=0.6cm
\begin{tabular}{c|c|ccc|ccc}
\hline
Method&	Venue&	Architecture&	Sea&	Offshore water&	OS*&	UNK&	HOS\\
\hline
OSBP \cite{19OSBPSaito10.1007/978-3-030-01228-1_10}&	ECCV'18&	65.8&	56.3&	73.4&	65.2&	15.5&	25.0\\
STA \cite{20STAliu8953906}&	CVPR'19&	6.4&	\textbf{100.0}&	52.4&	52.9&	68.9&	59.9\\
DAMC \cite{DAMC9165955}&	TMM'21&	67.5&	75.2&	65.3&	69.3&	15.8&	25.8\\
UADAL \cite{UADALjang2022unknown}&	NeurIPS'22&	74.2&	5.5&	82.6&	54.1&	57.0&	55.5\\
ANNA \cite{21ANNAli10203690}&	CVPR'23&	66.7&	\underline{99.8}&	88.7&	85.1&	20.0&	32.4\\
MTS \cite{MTS10292696}&	TCSVT'24& 34.2&  91.2&    \underline{91.4}&    72.3&    64.5&    68.1\\
WGDT \cite{22WGDTbi10919145}&	TGRS'25&	\textbf{83.7}&	99.6&	81.2&	\underline{88.2}&	\underline{90.2}&	\underline{89.2}\\
\rowcolor{blue!10}
SoDa$^2$&	\textbf{Ours}&	\underline{79.7}&	99.5&	\textbf{97.0}&	\textbf{92.1}&	\textbf{97.4}&	\textbf{94.7}\\
\hline
\end{tabular}
\label{tab3}
\end{table*}


\subsection{Ablation Study}
To further investigate the contribution of each component, ablation experiments are conducted on three cross-scene tasks. We first analyze the impact of different loss terms. The overall loss function of the proposed method consists of three components: the classification loss ${{\mathcal{L}}_{\text{cls}}}$, the spectral decoupled alignment loss $\mathcal{L}_{spe}$, and the spatial decoupled alignment loss $\mathcal{L}_{spa}$. To examine the effect of each loss component on model performance, we design the following comparison settings: using only the classification loss, using the classification loss combined with the spectral decoupled alignment loss, using the classification loss combined with the spatial decoupled alignment loss, and using all three loss terms, i.e., the complete model. The results of the loss-function ablation experiments are reported in Table \ref{tab4}.

\begin{table}
\centering
\caption{Results of loss function ablation experiments} 
\renewcommand\arraystretch{1.4}
\tabcolsep=0.28cm
\begin{tabular}{ccccccc}
\hline
Task & ${{\mathcal{L}}_{\text{cls}}}$ & $\mathcal{L}_{spe}$ & $\mathcal{L}_{spa}$ & OS* & UNK & HOS \\ 
\hline
\multirow{4}{*}{PU-PC} & \checkmark &   &   & 59.4 & 66.7 & 62.8 \\
~ & \checkmark & \checkmark &   & 58.1 & 84.2 & 68.8 \\ 
~ & \checkmark &   & \checkmark & 63.8 & 78.7 & 70.5 \\ 
~ & \checkmark & \checkmark & \checkmark & 70.5 & 94.3 & 80.6 \\ 
\hline
\multirow{4}{*}{HU13-HU18} & \checkmark & ~ & ~ & 51.2 & 31.4 & 39.0\\ 
~ & \checkmark & \checkmark & ~ & 65.8 & 30.4 & 41.6  \\ 
~ & \checkmark & ~ & \checkmark & 56.9 & 44.8 & 50.1 \\ 
~ & \checkmark & \checkmark & \checkmark & 59.0 & 72.3 & 65.0 \\ 
\hline
\multirow{4}{*}{ZY-GF} & \checkmark & ~ & ~ & 58.3 & 47.1 & 52.1 \\ 
~ & \checkmark & \checkmark & ~ & 63.1 & 52.0 & 57.0 \\ 
~ & \checkmark & ~ & \checkmark & 69.0 & 88.8 & 77.7\\ 
~ & \checkmark & \checkmark & \checkmark & 92.1 & 97.4 & 94.7\\ 
\hline
\end{tabular}
\label{tab4}
\end{table}

The experimental results indicate that, across three cross-scene HSI tasks, the proposed SoDa$^2$ model consistently achieves the highest HOS values. This demonstrates that the designed loss terms can effectively work in a complementary manner to enhance the overall category recognition capability in open-set cross-scene classification. When only the classification loss is employed, the model yields the lowest HOS, with particularly poor performance in unknown-class recognition. This is mainly attributed to the distribution discrepancy between the source and target domains, which leads to domain shift and severely limits the model’s generalization ability to unknown samples. After incorporating the spectral decoupled alignment loss, both UNK and HOS exhibit stable improvements on the PU–PC and ZY–GF tasks. This observation suggests that aligning cross-domain spectral feature distributions facilitates the learning of more domain-invariant representations, thereby enhancing the discriminability of unknown classes. Similarly, adding the spatial decoupled alignment loss separately also improves model performance. For example, in the HU13-HU18 tasks, adding the spatial decoupled alignment loss significantly improves OS*, indicating that spatial structure alignment helps improve classification accuracy for known classes. When the classification loss, spectral decoupled alignment loss, and spatial decoupled alignment loss are jointly optimized, all evaluation metrics reach their best performance across the three tasks. The synergistic optimization of these loss components substantially enhances the robustness and overall performance of the model in complex cross-scene open-set classification scenarios, thereby validating the rationality and necessity of the proposed loss-function design.

Subsequently, we modified the multimodal feature extraction module and compared two different strategies for combining feature fusion and attention mechanisms. The first strategy involves fusing spatial and spectral features first and then applying a spatial-channel attention mechanism. The second strategy, which is the proposed method in this paper, first enhances the two types of features separately using spatial-channel attention mechanisms, followed by weighted fusion. This experiment aims to validate the impact of the order of the attention mechanism and feature fusion operations on the efficiency of multimodal information integration and the final recognition performance. The experimental results are shown in Table \ref{tab5}.

\begin{table}
\centering
\caption{Ablation Experiment Results of Feature Extraction Module}
\renewcommand\arraystretch{1.4}
\tabcolsep=0.18cm
\begin{tabular}{ccccccc}
\hline
\multirow{2}{*}{Metric} & \multicolumn{2}{c}{PU-PC} & \multicolumn{2}{c}{HU13-HU18} & \multicolumn{2}{c}{ZY-GF}\\ 
 \cline{2-7} 
& SoDa$^2$\_F&	SoDa$^2$	&SoDa$^2$\_F&	SoDa$^2$	&SoDa$^2$\_F	&SoDa$^2$\\
\hline
OS*	&64.6	&70.5	&60.0	&59.0	&91.4	&92.1\\
UNK	&78.3	&94.3	&37.4	&72.3	&61.6	&97.4\\
HOS	&70.5	&80.7 	&46.1	&65.0	&73.5	&94.7\\

\hline
\end{tabular}
\label{tab5}
\end{table}

As can be seen from the experimental results in Table \ref{tab5}, in the three cross-scene HSI classification tasks, the proposed method SoDa$^2$ outperforms its variant SoDa$^2$\_F in terms of HOS index. That is, the method of performing attention mechanisms separately before fusion is better than the method of fusing first and then performing attention mechanism, which verifies the effectiveness of the multimodal feature extraction module structure adopted.

\subsection{Parameter Analysis}
To further investigate the influence of key parameters on model performance, we conduct a series of parameter analysis experiments. 

(1) \textit{Analysis of parameter $K$}: We first analyze the effect of the number of Gaussian mixture components $K$ in the GMM. The experiments are carried out on three cross-scene tasks. We vary $K$ within the set \{2,3,4,5\} and evaluate the corresponding changes in OS*, UNK, and HOS on the PU–PC, HU13–HU18, and ZY–GF tasks. The experimental results are summarized in Table \ref{tab6}. The experimental results demonstrate that different numbers of Gaussian mixture components have distinct effects on model performance. As $K$ increases, both UNK and HOS exhibit a decreasing trend across all three tasks. When $K$=5, the HOS value reaches its lowest level, indicating that an excessively large number of mixture components tends to bias the model toward classifying most samples as known classes. This behavior degrades the model’s generalization ability to unknown samples and consequently weakens its discriminative capability for unknown categories. Therefore, to achieve a favorable balance between known- and unknown-class recognition, we set $K$=2 in all experiments.

\begin{table}
\centering
\caption{Experimental results with different numbers of $K$} 
\renewcommand\arraystretch{1.4}
\tabcolsep=0.35cm
\begin{tabular}{cccccc}
\hline
\multirow{2}{*}{Task}  & \multirow{2}{*}{Metric} & \multicolumn{4}{c}{$K$}\\ 
 \cline{3-6} 
 & & 2&3&4&5\\
\hline
\multirow{3}{*}{PU-PC} & OS*&	70.5	&64.7&	68.1&	72.3\\
 & UNK	&94.3	&71.4	&50.5	&28.3\\
 & HOS	&80.6	&67.9	&58.0	&40.7\\
\hline
\multirow{3}{*}{HU13-HU18} & OS*&	59.0	&56.0	&57.5	&70.5\\
 & UNK	&72.3	&38.0	&26.4	&15.6\\
 & HOS	&65.0	&45.3	&36.2	&25.6\\
\hline
\multirow{3}{*}{ZY-GF} & OS*&	92.1	&87.4	&97.8	&96.3\\
 & UNK	&97.4	&80.8	&66.1	&51.7\\
 & HOS	&94.7	&84.0	&78.9	&67.2\\
\hline
\end{tabular}
\label{tab6}
\end{table}

(2) \textit{Analysis of parameter $\alpha$}: To investigate the influence of the decoupled alignment loss coefficient on model performance, we conduct a parameter sensitivity analysis on the PU–PC dataset with different values of $\alpha$. Specifically, $\alpha$ is varied within the range \{1,2,3,4,5,6,7,8,9,10,11,12\}. The experimental results are reported in Table \ref{tab7}, including the three evaluation metrics OS*, UNK, and HOS. From the results, it can be observed that as $\alpha$ increases, the UNK metric shows an upward trend, indicating that strengthening the decoupled alignment loss helps the model better recognize unknown classes. In contrast, the OS* metric remains relatively stable across different values of $\alpha$, exhibiting only minor fluctuations. With respect to the HOS metric, a gradual improvement is observed as $\alpha$ increases, reflecting an overall enhancement in balanced performance. The HOS value reaches its maximum when $\alpha$ = 10, corresponding to the best overall model performance. When $\alpha$ is further increased to 11 and 12, HOS slightly decreases, suggesting that an excessively large domain adaptation weight may suppress the overall performance. Therefore, we set $\alpha$ = 10 in all experiments.

\begin{table}[t]
\centering
\caption{Experimental results with different numbers of $\alpha$} 
\renewcommand\arraystretch{1.4}
\tabcolsep=0.3cm
\begin{tabular}{cccc|cccc}
\hline
$\alpha$	&OS*	&UNK	&HOS	&$\alpha$	&OS*	&UNK	&HOS\\
\hline
1	&69.8	&75.8	&72.6	&7	&65.7	&93.3	&77.1\\
2	&65.0	&84.7	&73.6	&8	&65.2	&95.5	&77.5\\
3	&60.9	&91.0	&73.0	&9	&66.5	&95.9	&78.6\\
4	&68.1	&81.0	&74.0   &10	&70.5	&94.3	&80.6\\
5	&68.1	&82.8	&74.7   &11	&67.9	&89.9	&77.4\\
6	&63.3	&92.4	&75.1   &12	&67.2	&84.3	&74.8\\
\hline
\end{tabular}
\label{tab7}
\end{table}

\section{CONCLUSION}
\label{CONCLUSION}
This paper proposes a single-stage open-set domain adaptation method with decoupled alignment for cross-scene HSI classification. A contribution-aware dual-modality feature extraction module is designed to jointly capture spectral and spatial contribution-aware features of HSI. To learn domain-invariant representations, a decoupled alignment module is proposed. MMD is employed to separately minimize the distribution discrepancies between the source and target domains in both spectral and spatial feature spaces. An open-set recognition module is further designed, in which unconstrained intrinsic features extracted from the target domain are compared with MMD-aligned domain-invariant features through a feature consistency measurement. Based on the resulting consistency scores, the GMM is constructed to effectively distinguish known and unknown classes in the target domain. Extensive experiments conducted on three groups of HSI datasets demonstrate that the proposed SoDa$^2$ method achieves superior performance in terms of classification accuracy for both known and unknown classes. In addition, comprehensive ablation studies and parameter analyses are performed to investigate the contributions of individual loss terms, network components, and key parameters, further confirming the effectiveness of the proposed model.

\bibliographystyle{IEEEtran}
\bibliography{ref}

\begin{IEEEbiography}
[{\includegraphics[width=1in,height=1.25in,clip,keepaspectratio]{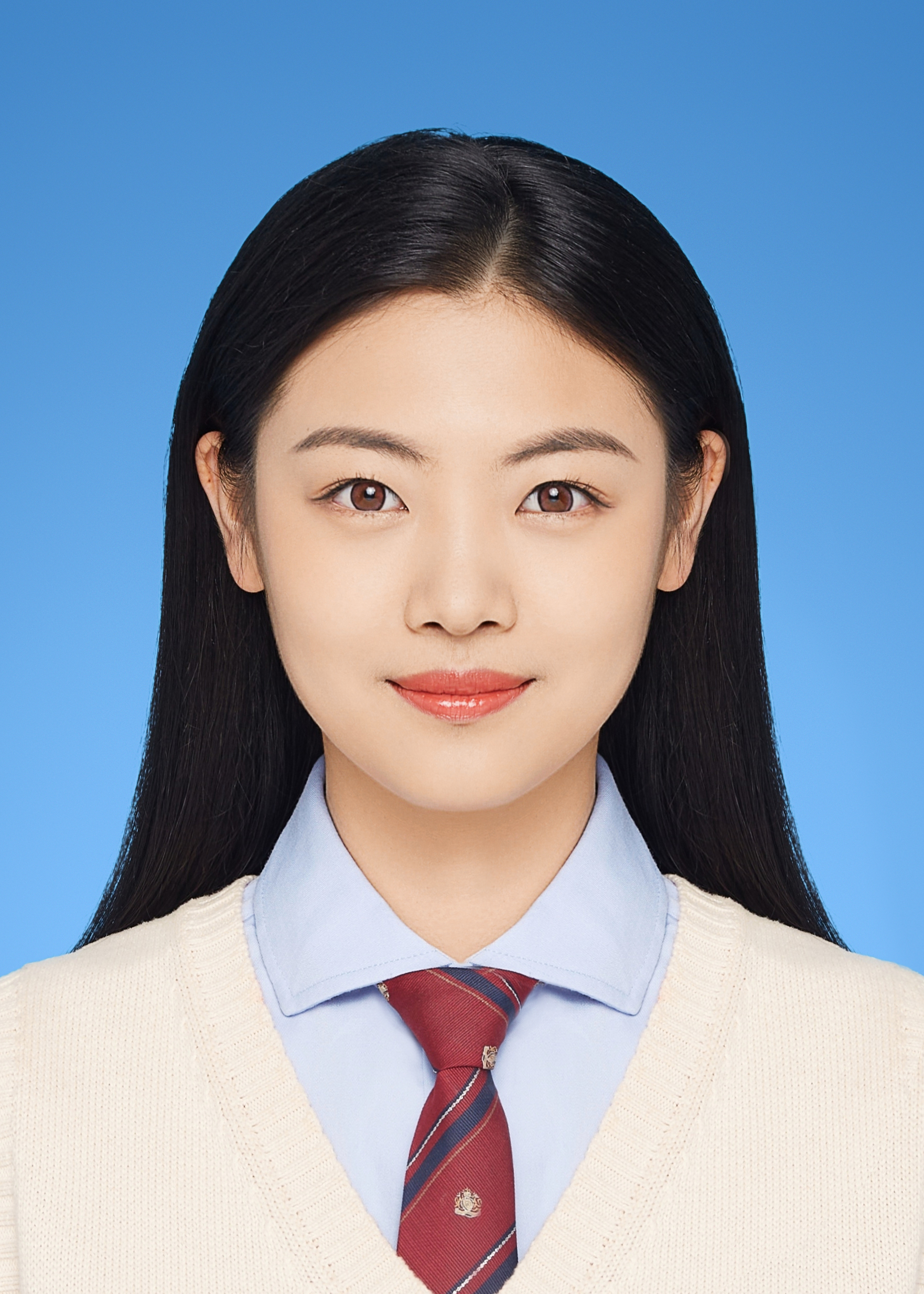}}]{Yiwen Liu} received the B.E. degree in Automation and the M.E. degree in Control Science and Engineering, Taiyuan University of Technology, Taiyuan, China, in 2021 and 2025, respectively. She is currently pursuing her Ph.D. degree in Intelligent Science and Technology with Nankai University, Tianjin, China. Her research interests include hyperspectral image classification, deep learning, and transfer learning.
\end{IEEEbiography}

\begin{IEEEbiography}
[{\includegraphics[width=1in,height=1.25in,clip,keepaspectratio]{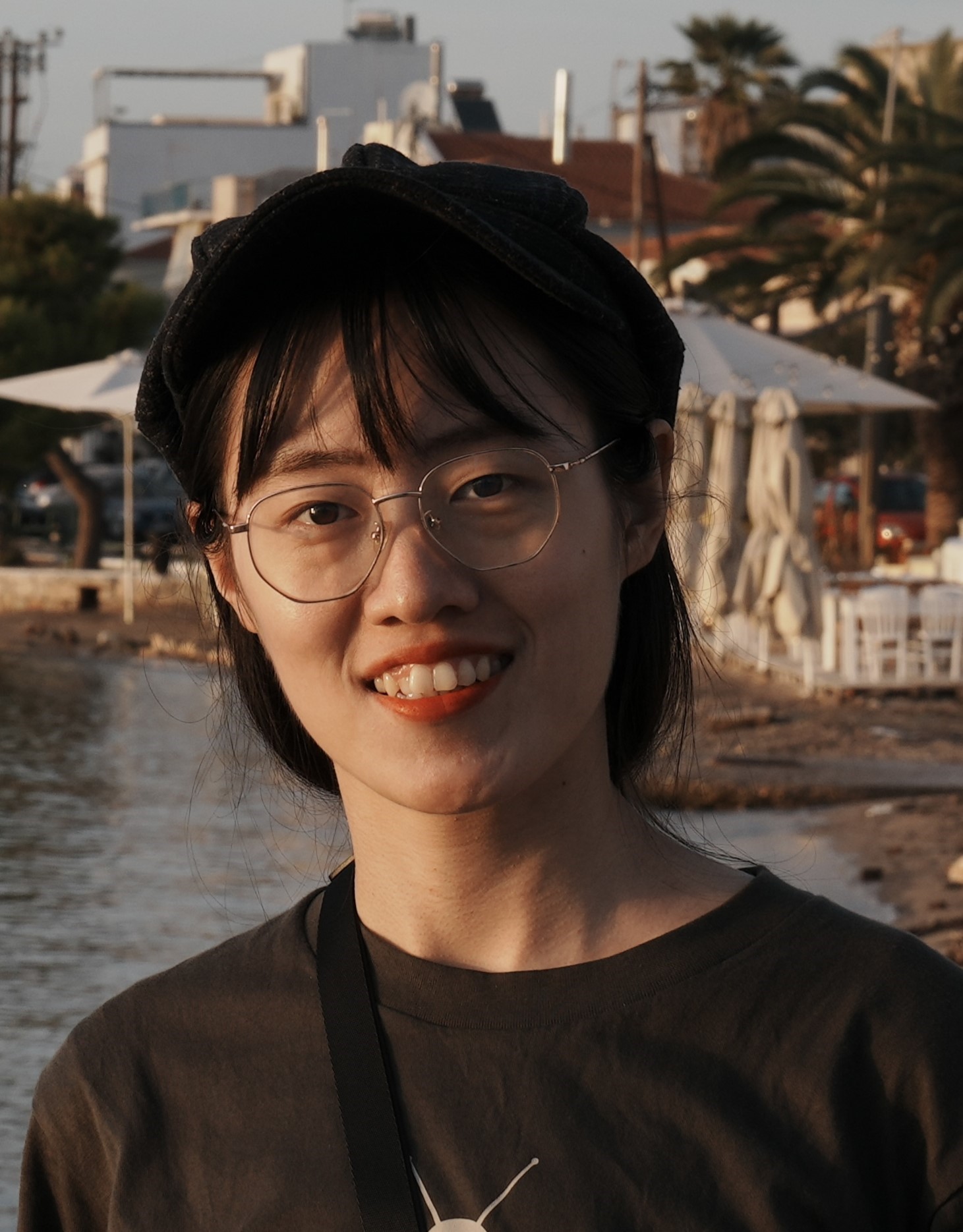}}]{Minghua Wang} (Member, IEEE) received the B.S. degree from the School of automation and the Ph.D. degree in Control Science and Engineering, Harbin Institute of Technology (HIT), Harbin, China, in 2016 and 2021, respectively. She was also a visiting Ph.D. student at the Univ. Grenoble Alpes, CNRS, Grenoble INP, GIPSA-lab, Grenoble, France (2019-2020). She worked with the Aerospace Information Research Institute, Chinese Academy of Sciences (CAS) (2021-2023). She is currently an Associate Professor at the College of Artificial Intelligence of Nankai University. Her research interests include remote sensing image processing, noise removal, anomaly detection, machine learning, and deep learning.
\end{IEEEbiography}

\begin{IEEEbiography}
[{\includegraphics[width=1in,height=1.25in,clip,keepaspectratio]{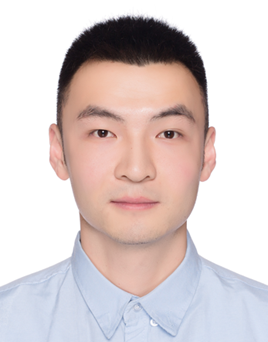}}]{Jing Yao} (Member, IEEE) received the Ph.D. degree in Mathematics from Xi’an Jiaotong University, Xi’an, China, in 2021. From 2019 to 2020, he was a visiting student at the Signal Processing in Earth Observation (SiPEO), Technical University of Munich (TUM), Munich, Germany, and the Remote Sensing Technology Institute (IMF), German Aerospace Center (DLR), Oberpfaffenhofen, Germany. Since 2021, he has been an Assistant Professor with the Aerospace Information Research Institute, Chinese Academy of Sciences, Beijing, China. His research interests include artificial intelligence for hyperspectral imaging, vision and language model for multimodal remote sensing. He was a recipient of the IEEE Geoscience and Remote Sensing Society Highest Impact Paper Award in 2025. He has been included in the Stanford's List of World's Top 2\% Scientists since 2024.
\end{IEEEbiography}

\begin{IEEEbiography}
[{\includegraphics[width=1in,height=1.25in,clip,keepaspectratio]{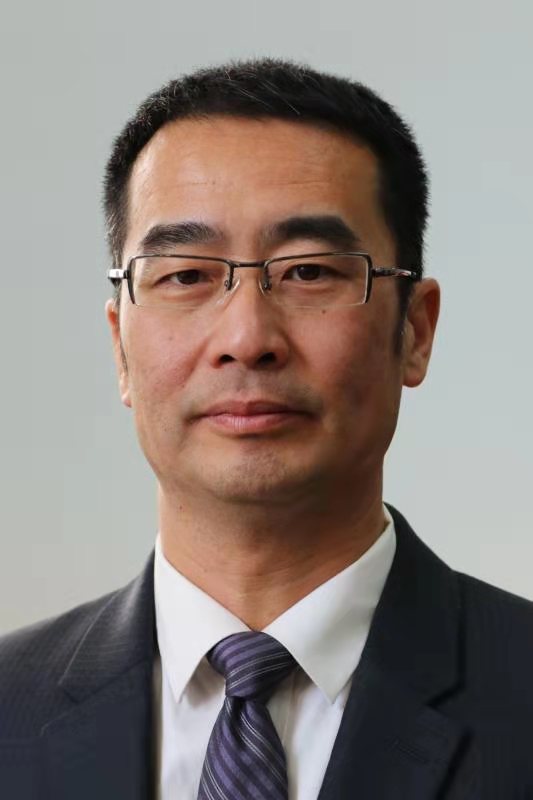}}]{Xin Zhao} (Member, IEEE) received the B.S. degree from Nankai University, Tianjin, China, in 1991, the M.S. degree from the Shenyang Institute of Automation, CAS, Shenyang, China, in 1994, and the Ph.D. degree from Nankai University in 1997, all in control theory and control engineering.

He joined the faculty at Nankai University in 1997, where he is currently a Professor and the Vice Dean of the College of Artificial Intelligence, Nankai University, Tianjin, China. He is also with Institute of Intelligence Technology and Robotic Systems, Shenzhen Research Institute of Nankai University, Shenzhen, China. His research interests are in micromanipulator, microsystems, and mathematical biology.
\end{IEEEbiography}

\begin{IEEEbiography}
[{\includegraphics[width=1in,height=1.25in,clip,keepaspectratio]{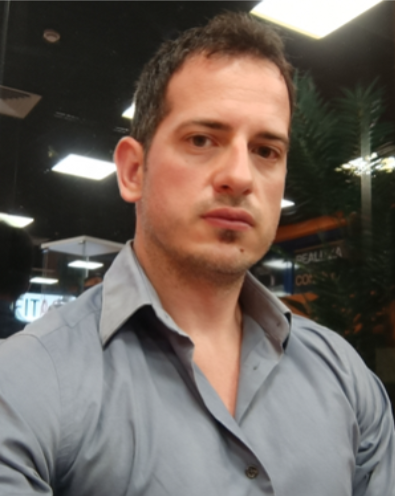}}]{Gemine Vivone} (Senior Member, IEEE) received the B.Sc. (summa cum laude), the M.Sc. (summa cum laude), and Ph.D. (Hons.) degrees in information engineering from the University of Salerno, Salerno, Italy, in 2008, 2011, and 2014, respectively. He is currently a Senior Researcher with the National Research Council, Tito Scalo, Italy. His main research interests focus on image fusion, statistical signal processing, deep learning, and classification and tracking of remotely sensed images. 

He received the IEEE GRSS Early Career Award in 2021, the Symposium Best Paper Award at the IEEE International Geoscience and Remote Sensing Symposium (IGARSS) in 2015, and the Best Reviewer Award of IEEE TRANSACTIONS ON GEOSCIENCE AND REMOTE SENSING in 2017. He is also the Editor-in-Chief of IEEE GEOSCIENCE AND REMOTE SENSING eNewsletter, an Area Editor of Elsevier Information Fusion, and an Associate Editor of IEEE TRANSACTIONS ON GEOSCIENCE AND REMOTE SENSING, IEEE JOURNAL OF SELECTED TOPICS IN APPLIED EARTH OBSERVATIONS AND REMOTE SENSING, and IEEE GEOSCIENCE AND REMOTE SENSING LETTERS. 
He is also an Advisory Board Member for SPRS Journal of Photogrammetry and Remote Sensing and an Editorial Board Member of MDPI Remote Sensing. He is listed in the World’s Top 2\% Scientists by Stanford University.

\end{IEEEbiography}

\end{document}